\documentclass[5p,times,twocolumn]{elsarticle}
\usepackage{amssymb}
\journal{Elsevier journal}
\date{January 20, 2023}

\begin{document}

\begin{frontmatter}

\title{Simple method for detecting sleep episodes in rats ECoG using machine learning}

\author[inst1]{Konstantin Sergeev}
\author[inst2,inst3,inst1]{Anastasiya Runnova}
\author[inst1,inst3,inst2]{Maxim Zhuravlev}
\author[inst4]{Evgenia Sitnikova}
\author[inst4]{Elizaveta Rutskova}
\author[inst4]{Kirill Smirnov}
\author[inst1]{Andrei Slepnev}
\author[inst1]{Nadezhda Semenova}
\ead{semenovni@sgu.ru}

\affiliation[inst1]{organization={Saratov State University},
            addressline={Astrakhanskaya str., 83}, 
            city={Saratov},
            postcode={410012}, 
            country={Russia}}           
\affiliation[inst2]{organization={Saratov State Medical University},
            addressline={B. Kazachaya Str., 112}, 
            city={Saratov},
            postcode={410012}, 
            country={Russia}}
\affiliation[inst3]{organization={National Medical Research Center for Therapy and Preventive Medicine},
            addressline={10(3) Petroverigsky Pereulok}, 
            city={Moscow},
            postcode={101990}, 
            country={Russia}}

\affiliation[inst4]{organization={Institute of Higher Nervous Activity and Neurophysiology of Russian Academy of Sciences},
           addressline={Butlerova street, 5A}, 
            city={Moscow},
            postcode={117485}, 
            country={Russia}}

\begin{abstract}
In this paper we propose a new method for the automatic recognition of the state of behavioral sleep (BS) and waking state (WS) in freely moving rats using their electrocorticographic (ECoG) data. Three-channels ECoG signals were recorded from frontal left, frontal right and occipital right cortical areas. We employed a simple artificial neural network (ANN), in which the mean values and standard deviations of ECoG signals from two or three channels were used as inputs for the ANN. Results of wavelet-based recognition of BS/WS in the same data were used to train the ANN and evaluate correctness of our classifier. We tested different combinations of ECoG channels for detecting BS/WS. 

Our results showed that the accuracy of ANN classification did not depend on ECoG-channel. For any ECoG-channel, networks were trained on one rat and applied to another rat with an accuracy of at least 80~\%. Itis important that we used a very simple network topology to achieve a relatively high accuracy of classification. Our classifier was based on a simple linear combination of input signals with some weights, and these weights could be replaced by the averaged weights of all trained ANNs without decreases in classification accuracy. In all, we introduce a new sleep recognition method that does not require additional network training. It is enough to know the coefficients and the equations suggested in this paper. The proposed method showed very fast performance and simple computations, therefore it could be used in real time experiments. It might be of high demand in preclinical studies in rodents that require vigilance control or monitoring of sleep-wake patterns.

\end{abstract}




\end{frontmatter}


\section*{Introduction}


The automatic recognition of various physiological states of laboratory animals is of great importance due to the progress in a framework of translational technologies of neuroscience and medicine~\cite{grandner2021translational,baran2022emerging}.
Multiparametric studies of pharmacological agents' effects in genetically homogeneous animals allowed breakthroughs in treatment of grave neurological disorders such as epilepsy and various neurodegenerative diseases~\cite{yang2018pathogenesis,khambhati2021long,fisher2019mouse}. Long-term physiological/behavioral parameters’ monitoring is often an integral part in such studies. For example, recent research including analysis of long-term recordings in WAG/Rij rats (absence epilepsy animal model) allowed to clarify different aspects of seizure activity development with age, gender and/or possibilities and impacts of pharmacotherapy~\cite{karadenizli2021age,lazarini2021absence,aygun2021exendin, knowles2022maladaptive}.

Various sleep studies have demonstrated that sleep structure is an important factor influencing central nervous system state~\cite{fink2018autonomic}, cognitive functions~\cite{mcsorley2019associations}, and other body systems functioning~\cite{seravalle2018sympathetic,cuddapah2019regulation}. In particular, discontinuous sleep, sleep fragmentation, and brief arousals often have a negative impact on health conditions. The nature of arousals during sleep has been reviewed in Ref.~\cite{halasz2004nature}, while the relationship between arousals and epilepsy has been thoroughly examined in P.~Hal{\'a}sz et al~\cite{parrino2006cap, halasz2013role,halasz2020sleep}. 

Sleeping brain studies usually require complex analysis of electroencephalographic (EEG), myographic (EMG), and also electrooculographic activities (when dealing with human sleep) that together form so-called polysomnogram (PSG)~\cite{sugi2009automatic,chapotot2010automated,van1984eeg}. PSG-based sleep scoring is the golden standard for determining states of sleep/wake cycle in the majority of species. At the same time, manual processing of even two or three days long EEG/PSG recordings, and sometimes additional video recordings, significantly complicates neurophysiological studies and increases their costs. There is a high need for reliable, precise, and accurate measurements of sleep parameters. 

A huge number of methods have been developed to detect specific normal sleep EEG oscillations and pathological EEG patterns, such as spike-wave discharges that tend to appear during drowsiness and sleep spindles (e.~g.,~\cite{sitnikova2014time,sitnikova2009sleep,sitnikova2016rhythmic}). While a certain number of methods for automatic sleep scoring in laboratory animals were introduced, the problem of automatic sleep scoring  has not been solved yet. EEG-based sleep detection and sleep staging methods are mainly based on time–frequency domain, amplitude, nonlinearity and other features~\cite{STUDLER2022, Bajaj2013, Acharya2015, MAHVASHMOHAMMADI2016,bastianini2014scoprism}. In particular, in our previous papers, we have demonstrated the possibilities of continuous wavelet transform (CWT) for detecting the physiological state of behavioral sleep and wakefulness in rats~\cite{runnova2021, runnova2022_1}. In these papers, the suggested method has shown high accuracy compared with an expert’s scoring. Nevertheless, both CWT and other methods of time-frequency processing PSG signals require significant numerical resources especially in case of long-duration recording. For these reasons, methods of PSG-data analysis using machine learning~(ML) have become widespread recently~\cite{You2022, Noorlag2022, Hussain2022, Wang2022}. 

In this paper we suggest a method allowing real-time sleep recognition without complex calculations or complex neural network types. In comparison to~\cite{KUMAR2022} where tens of characteristics has been extracted, here we use
only two electrocorticographic (ECoG) signal features namely mean value and standard
deviation, while our network consists of only one computational neuron. The use of this simple network not only eases the calculations but also clearly shows how mean and standard deviation of ECoG signals are related to switches of animals' state. Finally, we obtain one simple mathematical expression describing the sleep/wakefulness states of a laboratory rat.

\section{Materials and methods}
\label{sec:mat_met}
\subsection{Animals and ECoG recordings}
\label{sec:animals_ECoG}

Experiments in rats were conducted at the Institute of Higher Nervous Activity and Neurophysiology RAS (Moscow, Russia) according to EU Directive 2010/63/EU for animal experiments and approved by  institution’s animal ethics committee. Rats were kept in standard conditions with a natural light/dark cycle and had free access to rat chow and tap water. 
Eleven WAG/Rij male rats were used.

\begin{figure}[h]
\includegraphics[width=\linewidth]{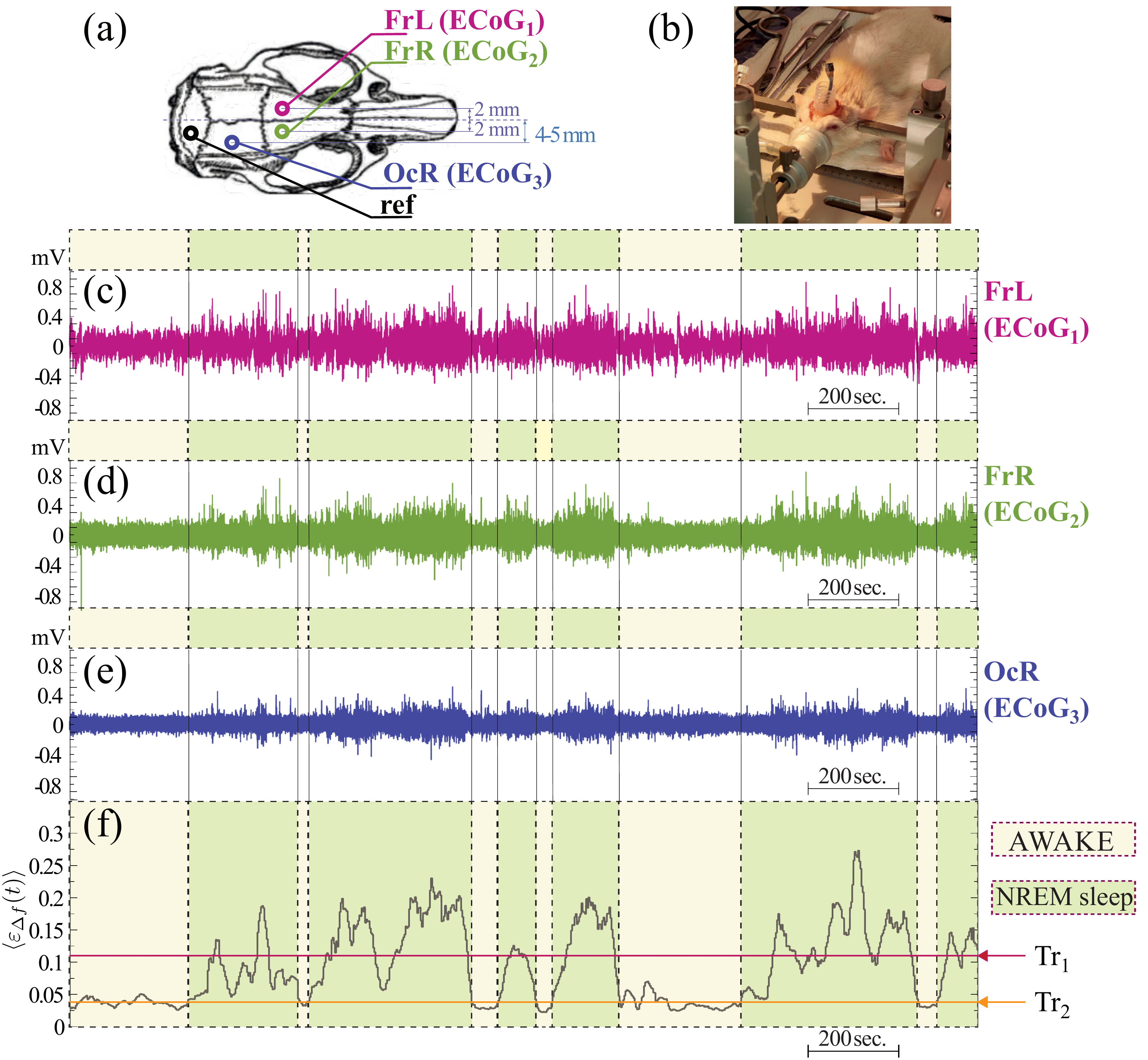}
\caption{(a) -- Scheme of electrodes location; (b) -- photograph of the electrode implantation surgery; (c), (d), (e) -- signal fragments ${\rm{ECoG}}_{1,2,3}(t)$ registered in left and right frontal and occipital canals, respectively; (f) -- a fragment of the characteristic $\mathbf{\Delta f}$ calculated from the given signals. The markup of WS/BS states is indicated by green and yellow colors.}
\label{fig:ECoG-CWT}
\end{figure}

All rats were implanted with screw electrodes for epidural ECoG recording (shaft length $= 2.0$~mm, head diameter $= 2.0$~mm, shaft diameter $= 0.8$~mm) over symmetrical right and left frontal cortex (AP $+2$~mm and L $\pm 2$~mm, FrR and FrL) and occipital (AP $-6$~mm and L $4$~mm, OcR). All coordinates are given relative to the bregma. The reference screw electrode was placed over the right cerebellum. 
Further, we used the notations ``$\rm{ECoG}_1$'' and ``$\rm{ECoG}_2$'' for frontal signals, and ``$\rm{ECoG}_3$'' for  occipital signal, Fig.~\ref{fig:ECoG-CWT}(a).

Stereotactic surgery was performed under sevoflurane anesthesia induced by using a nose cone (500~mL of 100\% O2 with sevoflurane $5\% \pm 0.4\%$).  Electrodes were permanently fixed to the skull with a methyl methacrylate monomer. The photograph of the surgery process is shown in Fig.~\ref{fig:ECoG-CWT}(b). Immediately after the surgery, rats received intramuscular injections of metamizole (FSSCI Microgen, Russia, 25~mg/kg) for pain relief and then were housed in individual cages in order to prevent damages to electrode connectors. Finally, they had been being left for recover within at least ten days before ECoG recording. 

After the surgery, rats were allowed to recover for a minimum of ten days. ECoG recordings were done in rats freely moved in plexiglas cages (25 x 60 x 60~cm) under 12:12~h light/dark cycle (light on at 8~a.~m.). ECoG signals fed into a multichannel amplifier (PowerLab 4/35, LabChart 8.0 software, ADInstruments, Sydney, Australia) via a swivel contact, band-pass filtered between $[0.5;200]$~Hz, digitized with 400~samples/second per channel and stored in the hard disk. The full-length ECoG recording session lasted approximately 24~h (avg 22.2~h, min 20.0~h, max 26.6~h). Immediately prior to the full-length ECoG recording, video-ECoG was recorded for 1~h using video camera Genius eFace 1325R and video capture module for LabChart.

Wavelet-based ECoG analysis was performed during the dark phase.  
This implies that disturbances of sleep control mechanisms were well pronounced. Analysis was done during two 3~h periods of light/dark cycle: at the beginning of dark phase 21:00~--~23:59 (``beginning of the dark phase'') and at the end of dark phase 3:00~--~5:59 (``end of the dark phase'').


 
One-hour video-ECoG recordings were visually inspected in all subjects at the age of 9 months. 

ECoG signals $\left\{{\rm{ECoG}}_1(t), {\rm{ECoG}}_2(t), {\rm{ECoG}}_3(t) \right\}$ with duration $T$ were recorded with a sampling frequency~$(1/N)=400$, i.~e., the recording of duration 1~s contains $N$~values. The examples of these ECoG recordings are shown in Fig.~\ref{fig:ECoG-CWT}(c--e).

\subsection{Training markup based on wavelet approach}\label{sec:markup}
Behavioral sleep~(BS) and waking state (WS) were detected using a wavelet-based algorithm, introduced previously in \textit{Runnova, et al.}~(2021)~\cite{runnova2021, runnova2022_1}. Figure~\ref{fig:ECoG-CWT}~(f) shows the BS/WS markup based on this wavelet approach with different thresholds. This technique is discussed in more detail in Supplementary materials (Sect. SM.1). The described method was originally developed to detect periods of deep NREM sleep. The problem of finding difference between periods of REM sleep and waking states using this method has not been discussed.

\subsection{Machine learning approach}
\label{sec:ML_approach}

The proposed method for recognising behavioral sleep (BS) and waking state (WS) was based on machine learning approach, namely an artificial neural network (ANN). Usually raw signal data are not used as an input of ANN. Instead, data-scientists extract different ``features'' from analysed signals~\cite{KUMAR2022,majkowski2018implementation}, such as different statistical characteristics, entropy, fractal, time-frequency,  spatio-temporal and other characteristics. The choice of the correct set of characteristics largely determines the success of ANN application and final accuracy. In the case of sleep recognition based on ECoG signals, these features can be averages, variations and deviations from the mean~\cite{hassan2017, hsu2013, Ronzhina2012, zhu2014}. In our previous studies of ECoGs with ANN, we used signal-to-noise ratio including mean and standard deviation of ECoG signal~\cite{our_BBB,Glushkovskaya2023}. It follows that mean~$\mu({\rm{ECoG}})$ and standard deviation~$\sigma({\rm{ECoG}})$ can be very appropriate for the primary task of this paper. The motivation of this choice is additionally underlined in Supplementary materials (Sect. SM.2)

In this paper, the ANN was constructed and trained using an open-source deep learning API Keras~\cite{keras}. The training set was prepared and normalized using standard Python tools~(TensorFlow, NumPy, pandas, etc.).

\subsubsection{Data preprocessing}
\label{sec:data_prep}

The recorded ECoG data in millivolts contain values between $-200$ and $200$ mV. 
 Before calculating the mean and standard deviation, we use the next normalization technique in order to overcome the problem with between-subject variability in ECoG amplitude
 First, we make the values only positive by $Y(t)=\mathrm{ECoG}_i(t)-\mathrm{min}(\mathrm{ECoG}_i)$. Next, we put them into the same range [0;1) rationing $Y(t)$ in accordance with
\begin{equation}
\mathrm{ECoG}^*_i(t)=\frac{Y(t)}{2\cdot \mathrm{mean}(Y)}.
\label{eq:normalization}
\end{equation} 

The averaged characteristics $\mu(ECoG)$ and $\sigma(ECoG)$ were computed during the fixed time interval. Its duration is an important parameter depending on both sampling frequency and signal specifics. The optimal averaging window duration was found empirically $\delta=10$ sec. Since the sampling frequency $f_d$ was 400 sample values per second, the averaging window contained $N=\delta f_d=4000$ sample values. Then the set of mean values can calculated from $\mathrm{ECoG}^*_i$ as
\begin{equation}
\mu_i(t_j) = \frac{1}{\delta f_d} \sum_{k=1}^{\delta f_d}{\mathrm{ECoG}^*_i\big(t_j-\frac{k}{f_d}\big)},
\label{eq:mu}
\end{equation}
where $i=1,2,3$ is the ECoG channel, and $\mu_i(t_j)$ is its mean value averaged over the time interval from $(t_j-\delta)$ to $t_j$. The choice of this interval is discussed in more detail in Supplementary materials (Sect.~SM.5).

The same technique was applied for calculating the standard deviation:
\begin{equation}
\sigma_i (t_j ) = \frac{1}{\delta f_d-1} \sum_{k=1}^{\delta f_d}{\Big|\mathrm{ECoG}^*_i\big(t_j-\frac{k}{f_d}\big)-\mu_i(t_j)\Big|^2}.
\label{eq:sigma}
\end{equation}

\begin{figure*}[t!]
\center{\includegraphics[width=1\linewidth,keepaspectratio]{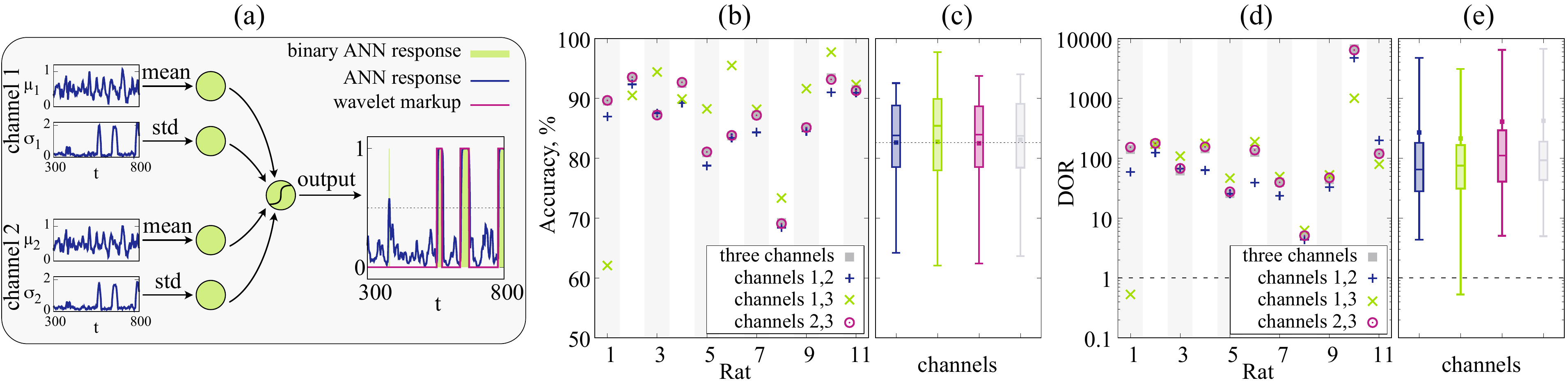}}
\caption{The first panel (a) is the scheme of considered ANN, where left panels correspond to examples of realizations for mean $\mu_i(t_j)$ and standard deviation $\sigma_i(t_j)$ calculated on ECoG implementations of channels 1 and 2 with equations (\ref{eq:mu}, \ref{eq:sigma}). The right part of panel (a) shows the corresponding realization of ANN responses, which was trained on channels 1,2 of animal \#1 and applied to the data of animal \#11. The pink line corresponds to the wavelet-based markup, wile the blue and green colors represent the ANN response and its binary transformation with the threshold 0.5, respectively. Panels (b) and (d) show the accuracies and DOR characteristics of ANNs trained on the data of animal \#1 and applied to the rest animals, respectively. Four ANNs were trained in this manner for different combinations of channels: 1\&2 (blue), 1\&3 (green), 2\&3 (pink) and all three channels (gray). Panel (c) and (e) represent the averaged statistics of all accuracies and DOR values of all ANNs trained for animals \#\#1,2,4,7,8 and applied to the rest 10 animals shown by boxplots with the same color scheme as in (b,d). The dashed line is the averaged accuracy level for channels ${\rm{ECoG}}_1(t)$ and ${\rm{ECoG}}_2(t)$. Parameters: $\delta=10$ sec, $\tau=1$ sec.}
\label{fig:scheme_accuracy_DOR}
\end{figure*}

In order to increase the number of testing and training examples and to make ANN responses more smooth, we calculated the mean and standard deviation in a shifting window, where window shift $\tau=1$ sec was less than the window size $\delta$. Then $t_j=j\cdot\tau+\delta$ in Eqs.~(\ref{eq:mu},\ref{eq:sigma}). Therefore, the original one-hour $\mathrm{ECoG}_i(t)$ implementation was transformed into two new implementations $\mu_i(t_j)$ and $\sigma_i(t_j)$ containing  $3590$ points. Their examples are given in the left part of Fig.~\ref{fig:scheme_accuracy_DOR}(a). 

The normalization procedure (\ref{eq:normalization}) was applied to data twice: 1 -- to initial ECoG data and 2 -- to $\mu_i(t_j)$ and $\sigma_i(t_j)$. The second normalization shifted the range of values into the interval $(0,1)$ which was necessary for correct ANN application (see Supplementary materials Sect.~SM.4).

\subsubsection{ANN structure}
\label{sec:ANN_struct}

One common use of machine learning methods is identifying the important features and how are they related. 
 Figure~\ref{fig:scheme_accuracy_DOR}(a) shows a scheme how ANN can get mean values and standard deviations from two channels as an input.
 As can be seen from Figure~\ref{fig:scheme_accuracy_DOR}(a), it has only four input neurons and one artificialneuron right after them.
 After training, the proposed simplified network will allow to replace the use of a neural network with a simple linear combination of input signals.
 This can potentially lead to faster classifier performance and the ability to work in real time.
 We have considered several combinations of ECoG channels.
 Our preliminary study indicated that a single-channel ECoG data were not sufficient for classification. 
 Therefore, further we used two or three ECoG channels. 
 
The ANN shown in Fig.~\ref{fig:scheme_accuracy_DOR}(a) is often referred to as a single-layer perceptron~\cite{Haykin}.
 It has four inputs receiving the mean values and standard deviations $\mu_1, \sigma_1$ and $\mu_2, \sigma_2$ calculated by channels 1 and 2. 
 ANN can have two more input signals $\mu_3, \sigma_3$ when using all three channels simultaneously $\left\{{\rm{ECoG}}_1(t), {\rm{ECoG}}_2(t), {\rm{ECoG}}_3(t) \right\}$.

The next green circle after these four inputs is the nonlinear artificial neuron with sigmoid activation function $f(x)=1/(1+e^{-x})$, commonly used in ANNs~\cite{Haykin}. 
 This activation function produces output real values from the range $(0;1)$. 
 Mathematically speaking, if the vector of input values is $\vec{x}_\mathrm{in}=(\mu_1,\sigma_1,\mu_2,\sigma_2)$, then the output of the network becomes:
\begin{equation}
y_\mathrm{out} = f(\vec{x}_\mathrm{in} \cdot \mathbf{W} +b),
\label{eq:simple_in_matrix}
\end{equation}
where $\mathbf{W}$ is the weight matrix connecting the input and output layers. Its size is $(4\times 1)$ for 2 channels and $(6\times 1)$ for 3 channels. The bias $b$ allows to shift the output values along the activation function.
 For training the network for BS/WS recognition we use the markup based on the wavelet approach. The values $\mathbf{W}$ and $b$ are varied to make the network response $y_\mathrm{out}$ as similar as possible to the originally known markup during the training of the network. 
 The network is trained to answer 0 if the input values are obtained in waking state (WS), while 1 corresponds to behavioral sleep (BS).
 Therefore, very complex wavelet-based method of WS/BS recognition described in Sect.~\ref{sec:markup}, can be compared with artificially obtained classifier based on the simple algebraic equation of interrelated characteristics:
\begin{equation}
y_\mathrm{out} = f(\mu_1 W_{\mu 1} + \sigma_1 W_{\sigma 1} + \mu_2 W_{\mu 2} + \sigma_2 W_{\sigma 2} + b),
\label{eq:simple_algebraic}
\end{equation}
where $\left\{ W_{\mu_1},W_{\sigma_1}, W_{\mu_2}, W_{\sigma_2} \right\}$ are the components of the weight matrix $\mathbf{W}$ of ANN given in Fig.~\ref{fig:scheme_accuracy_DOR}(a). These coefficients will be further called as \textit{weight coefficients}.

\subsubsection{ANN training}
\label{sec:ANN_train}

For training the ANN we used the free API Keras and programming language Python. 
 All trained networks used in this paper were trained in the same way with Adam optimization algorithm and binary crossentropy loss function. Such network training implied the use of initially marked up data based on the wavelet approach \citep{runnova2021} (see Sect.~\ref{sec:markup}).
 The ``ground-truth'' of this data will be discussed in more detail in Supplementary materials SM.2.

Each network was trained on means and standard deviations of continuous ECoG epochs extracted from 1-hour records lasting from 1000 to 2000 s. These epochs were selected in a way to include the equal length of BS and WS.
 The data used for one training were recorded from one of 11 animals. 
 Next, this trained ANN was used to analyse the data from the remaining animals. 
 
Further, we are going to focus on the results obtained for all three ECoG channels and different combinations of two channels. 
 The accuracy of detecting was calculated as a percentage and calculated as the ratio between correct answers and the total number of input data. 
 The training accuracy was at least 80\% across all rats. This number was slightly increased for some training datasets.

\section{Results}
\label{sec:results}
\subsection{BS/WS recognition based on frontal and occipital ECoGs}
The right part of Fig.~\ref{fig:scheme_accuracy_DOR}(a) shows the example of applying the trained ANN to ECoG implementation.
 The pink curve is the markup based on the wavelet approach.
 The ANN response is shown by the two curves: 1 -- pure response without any changes (blue); 2 -- binary form (0,1) with the threshold $0.5$ (green).
 This threshold value is fixed throughout the paper.

In order to analyse and compare the performance of ANNs trained on different combinations of ECoG channels, we will use two characteristics: accuracy and diagnostic odds ratio~(${\rm DOR}$)~\cite{Glas2003}. The accuracy is calculated as the ratio of true positive ($TP$) an true negative ($TN$) answers to the total number containing additional false positive ($FP$) and false negative ($FN$) ANN answers. The DOR depends on the same characteristics but in a different way:
\begin{equation}
A = \frac{TP + TN}{TP+TN+FP+FN}; \ \ \ \ \ 
{\rm DOR} = \frac{TP \cdot TN}{FP \cdot FN}.
\label{eq:DOR}
\end{equation}
The {\rm DOR} ranges from zero to infinity. If it is less than 1, then the network works in a wrong direction and its performance can be improved by inversion of the network's answers \cite{Glas2003}. The {\rm DOR} equal to 1 indicates that the network produces the same response regardless of the input data. If the {\rm DOR} is greater than unity, then the network is well trained and gives useful answers. In this case the higher {\rm DOR} values correspond to the better network performance. 

First, we considered neural networks that were trained on data recorded from only one of the animals.
 The part $t\in[1000;3000]$ sec. of ECoG implementations was used for training. The final training accuracy of ANN trained on two frontal channels of animal \#1 is $A^\mathrm{tr.1}_{12}=85.4\%$. The other combinations of channels give training accuracies $A^\mathrm{tr.1}_{13}=50\%$, $A^\mathrm{tr.1}_{23}=87.5\%$, $A^\mathrm{tr.1}_{123}=87.9\%$.
 These networks can be applied to BS/WS recognition of the rest animals, given the corresponding set of channels.
 Figure~\ref{fig:scheme_accuracy_DOR}(b) shows the accuracies for all four ANNs trained on a data of animal \# 1 and applied to other data.
 The color shows the different combinations of channels: 1 -- left and right frontal channels (blue color), 2 -- left frontal and occipital (green), 3 -- right frontal and occipital (pink), and 4 -- all three channels (gray).

As can be seen from Fig.~\ref{fig:scheme_accuracy_DOR}(b), the worst accuracy about 70\% is obtained for animal \#8, whereas the data of the rest animals provide the accuracies at least 80\%.

Similar results were obtained for other ANNs trained on animals \#\#2,4,7 and 8 having the maximal percentage of sleep during one-hour realizations. 
The degree of homogeneity of our results indicates a good repeatability of the proposed approach.

Figure~\ref{fig:scheme_accuracy_DOR}(c) shows the statistical analysis of accuracies for all five ANNs (trained on animals \#\#1,2,4,7,8) and applied to other 10 animals. The combinations of different channels are shown in the same color scheme as in panel (b). The mean accuracy of ANNs for frontal channels is 82.5\% indicated by the dashed line in Fig.~\ref{fig:scheme_accuracy_DOR}(c). The averaged accuracies 83.1 and 82.4\% were obtained for occipital channels combined with the left and right frontal channels, respectively. The averaged accuracy for all three channels is 82.9\%.

Figure~\ref{fig:scheme_accuracy_DOR}(d,e) was prepared in the same manner but for {\rm DOR} characteristics. The obtained {\rm DOR} values are significantly larger than unity indicating good network performance. Moreover, Fig.~\ref{fig:scheme_accuracy_DOR}(e) indicates, that the median and mean DOR values exceed 65 and 200, respectively. 

Thus, all four types of ANNs demonstrate similar accuracy and similar statistics inside the network. Therefore, it is of particular interest to uncover their different and common features, discussed in the next subsection.

\subsection{How the mean and standard deviation multipliers affect the BS/WS recognition}
\label{sec:results_universal_W}
During the training process, the weights of ANN are changing so that the network responses agrees with the training data. The ANNs considered in this paper have only one nonlinear neuron with four or six inputs. Therefore, it is very easy to understand which input signals are important and what is their combination. According to Eq.~(\ref{eq:simple_algebraic}), the ANNs output is the combination of inputs with corresponding multipliers $W_{\mu i}$ and $W_{\sigma i}$, where $i=1,2,3$.

\begin{figure}[t]
\center{\includegraphics[width=0.9\linewidth,keepaspectratio]{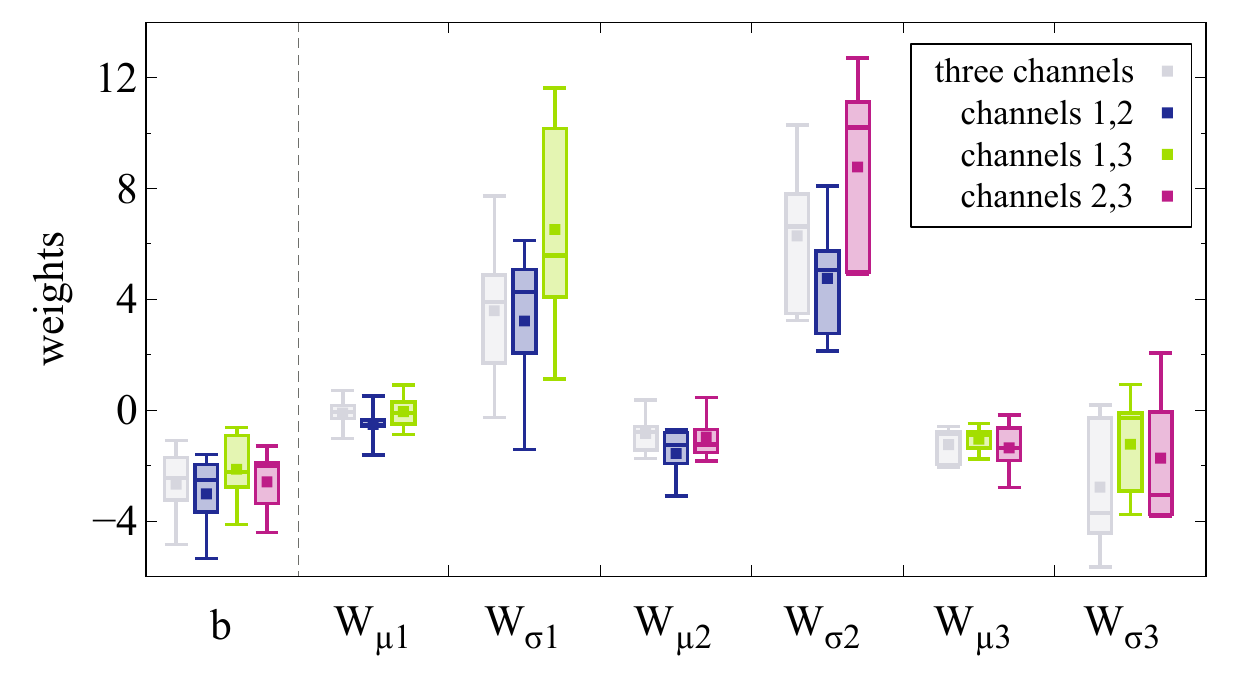}}
\caption{Comparison of biases $b$ and weights $W_{\mu i}, W_{\sigma i}$ of ANNs trained on different combinations of channels: 1\&2 (blue), 1\&3 (green), 2\&3 (pink) and all three channels (gray). The boxplots were prepared according the statistics of ANNs trained on the dataset of animals \#\#1,2,4,7,8.}
\label{fig:synaptics}
\end{figure}

The statistics of obtained weights is given in Fig.~\ref{fig:synaptics}. They were calculated according five ANNs trained on data from animals \#\#1,2,4,7,8 for each combination of channels. The color scheme of considered channels is the same as in Fig.~\ref{fig:scheme_accuracy_DOR}(b--e).

The statistical analysis of weight coefficients $W_{\mu i}$ and $W_{\sigma i}$ gives an unexpected result. Despite the fact that the weights were initially set randomly and the networks were trained for different animals, the statistics of the final weights for particular $W_{\mu i}$ and $W_{\sigma i}$ turned out to be similar, especially $W_{\mu i}$. The scatter of weight coefficients $W_{\sigma i}$ is larger. Nevertheless their values for particular $i$ are also fairly close. 

The above results indicate the generality of properties $\mu_i$ and $\sigma_i$ of the ECoG signals from the frontal and occipital parts of the cortex, namely:
\begin{itemize}

\item the synaptic multipliers $W_{\mu i}$ responsible for the mean value $\mu$ of signals from both frontal and occipital parts are negative and relatively small in magnitude;
\item the synaptic multipliers $W_{\sigma 1}$ and $W_{\sigma 2}$ responsible for the standard deviation $\sigma$ of signals from the frontal cortex are positive and relatively large in magnitude;
\item multipliers $W_{\sigma 3}$ responsible for the standard deviation $\sigma$ of signals from the occipital cortex are negative. And the same can be said about bias $b$ regardless of the channels being considered. 
\end{itemize}

In order to prove these assumptions, we have constructed the ANNs whose weight coefficients were averaged over all previously discussed ANNs corresponding to channels 1\&2, 1\&3, 2\&3 and all three channels. 

\begin{figure}[t]
\center{\includegraphics[width=0.9\linewidth,keepaspectratio]{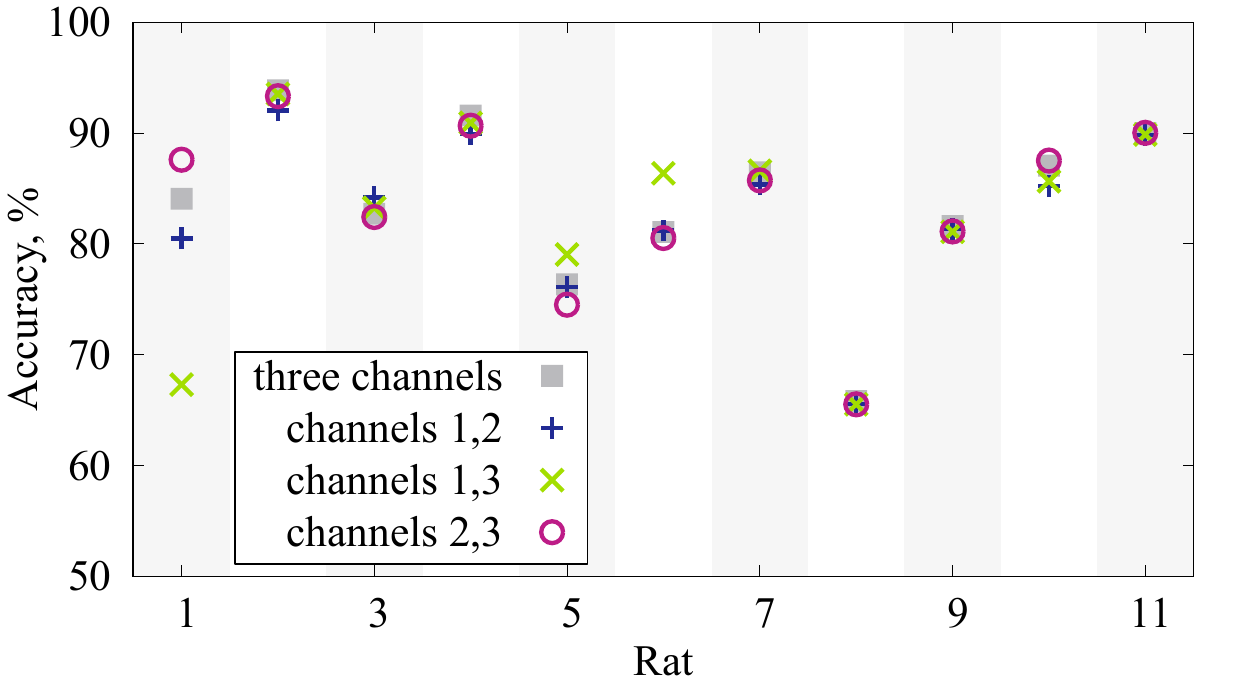}}
\caption{Accuracy of ANN with averaged wight coefficients applied to all 11 animals.}
\label{fig:Wk_acc}
\end{figure}

The weights of these averaged ANNs are given in Table~\ref{tab:averaged_weights}. They demonstrate a high level of BS/WS classification accuracy in all 11 animals (Fig.~\ref{fig:Wk_acc}). Therefore, using Eq.~(\ref{eq:simple_algebraic}) and suggested in Table~\ref{tab:averaged_weights} averaged weight coefficients, one can simply recognize BS/WS without training the networks.

\begin{table*}\begin{center}
\begin{tabular}{c | c | c | c | c | c | c | c || c}
\textbf{Channels} & $b$ & $W_{\mu 1}$ & $W_{\sigma 1}$ & $W_{\mu 2}$ & $W_{\sigma 2}$ & $W_{\mu 3}$ & $W_{\sigma 3}$ & \textbf{Averaged accuracy}  \\ \hline
 left and right frontal (1,2)     & -3.02 & -0.51 &  3.22 & -1.56 &  4.76 &  --   &  --   &  82.85\\
 left frontal \& occipital (1,3)  & -2.13 & -0.06 &  6.52 &  --   &  --   & -1.05 & -1.23 &  82.66	\\	
 right frontal \& occipital (2,3) & -2.59 &  --   &  --   & -0.97 &  8.78 & -1.36 & -1.73 &  83.55	\\
 all three (1,2,3)                & -2.67 & -0.10 &  3.59 & -0.84 &  6.29 & -1.24 & -2.78 &  83.70
\end{tabular}
\end{center}
\caption{Averaged weight coefficients $W_{\mu_i}, W_{\sigma_i}$, biases $b$ and averaged accuracy of BS/WS recognition using ANN with these weights.}
\label{tab:averaged_weights}
\end{table*}

\section{Discussion}
\label{sec:discuss}

\textit{Is an ANN necessary to solve BS/WS recognition problem?}

Started with a complex deep network analysing $\mu$ and $\sigma$ of three ECoG channels and giving an accuracy of about 80~--~90~\, we came to simplifying the network structure.
 It was surprising to find out that only $1$ neuron with the sigmoid activation function could do this task with the same accuracy.
 Because of this extreme simplicity of the network, its learning algorithm was just a method to get the weights $W_{\mu i}$, $W_{\sigma i}$ and bias $b$ such that Eq.~(\ref{eq:simple_algebraic}) produces the answer closest to the correct.
 
Moreover, in Sect.~\ref{sec:results_universal_W} we have considered a new ANN with averaged connection weights, which are averaged across weights $W_{\mu i}$, $W_{\sigma i}$ and $b$ of all previously trained ANNs.
 The accuracy of this artificially obtained ANN is comparable with other trained ANNs. 
 Therefore, now it is not necessary to train the network, the use of simple algebraic equation (\ref{eq:simple_algebraic}) with calculated averaged coefficients is enough.
 As an alternative to ANN, there are various methods of unsupervised learning such as clustering (regressions, k-means,  Kohonen map, etc).
 However, these methods are beyond the framework of this article and requires a separate study.

\section*{Conclusion}

We have developed an original automatic classification algorithm recognising sleep episodes on ECoG signals in rats.
 The suggested method is based on neural network including the preliminary calculation of simple statistical characteristics and further classification based on their linear combination.

Our approach requires at least two ECoG channels. 
 There is no exact requirements in which part of the cortex the electrodes should be placed. 
 Our calculations show that the use of different combinations of frontal left, frontal right and occipital signals lead to similar classifier accuracy. 
 The averaged accuracy of obtained networks was 80--90\% while the DOR characteristics were much larger than unity. 

Moreover, here we suggest very simple neural network to recognize BS/WS episodes. This network is so small that it can reveal a linear combination of mean values and standard deviations with certain coefficients. These coefficients are the result of averaging of weights from several trained networks for different animals. Using these averaged weight coefficients (Table~\ref{tab:averaged_weights}) in Eq.~(\ref{eq:simple_algebraic}), one can recognize sleep without training the network.

The proposed method allows to analyse the appearance of base sleep in almost real time. It can be very useful, for example, for development of laboratory devices for sleep deprivation. 

\section*{Data availability}
The data that support the findings of this study are available from the corresponding author upon reasonable request.

\section*{Acknowledgements}
The recognising sleep ANN was developed and improved in the framework Russian Science Foundation Grant No. 21-75-10088. Its application to data and comparison with other methods was partially supported by RF Government Grant No. 075-15-2022-1094. Experimental data were obtained within the state assignment of the Ministry of Education and Science of the Russian Federation to Institute of Higher Nervous Activity for 2021–2023.

%
%

\newpage
\textbf{SUPPLEMENTARY MATERIALS}

\setcounter{equation}{6}
\setcounter{figure}{4}
\setcounter{section}{0}

\section{Wavelet method of behavioral Sleep/Wake detection}
\label{sec:wavelet}

The continuous wavelet transformation~(CWT) $W_i(f, t)$ was calculated for each ECoG signal ${\rm{ECoG}}_i(t)$ in this complex form
\begin{equation} \label{eq:CWT}
W_i\left(f,t_0\right)=\sqrt{f}\cdot\int^{T}_{0}{\rm{ECoG}}_i\left(t\right)\cdot\Psi_{f,t_0}^*\left(\frac{t-t_0}{f}\right)\cdot dt,
\end{equation}
\noindent where $t_0$ specifies the wavelet location on the time axis, ``*'' denotes the complex conjugation, and $\Psi\left(f,t\right)$ is the base wavelet function. We use the standard Morlet wavelet with the scaling parameter $2\pi$:
\begin{equation} \label{eq:baseCWT}
\Psi_{f,t_0}\left(x\right)=\frac{1}{\sqrt[4]{\pi}}\cdot e^{-1/4}\cdot e^{\imath\cdot 2\pi\cdot x} e^{-\frac{x^2}{2}},
\end{equation}
\noindent where $\imath$ is an imaginary unit. 
Morlet wavelet is widely used for analysing the signals of biological objects, allowing to maintain the optimal ratio between the frequency and time resolution achieved with CWT~(\ref{eq:CWT})~\cite{Pavlov:2012_YFN_engl,Sitnikova2013_BrainResearch}. Each one-dimensional signal ${\rm{ECoG}}_i(t)$ allows estimating a two-dimensional wavelet surface as:
\begin{equation} \label{eq:baseCWT2}
W_i\left(f,t_0\right)=\left| W_i\left(f,t_0\right)\right| \cdot e^{\imath \cdot \Psi_{f,t_0}^*\left(\frac{t-t_0}{f}\right)}.
\end{equation}

This surface $W_i(f, t_0)$ characterizes the oscillatory activity for each frequency $f$ at any time $t_0$ for the initial signal ${\rm{ECoG}}_i(t)$. Next, we define the integrated energy distribution $\left\langle E^{\Delta f}_i(t_0) \right\rangle$ for a certain frequency range $\Delta f = [f_\mathrm{min}; f_\mathrm{max}]$ as:
\begin{equation} \label{eq:energy}
\left\langle E^{\Delta f}_i(t_0) \right\rangle=\int_{f_\mathrm{min}}^{f_\mathrm{max}}\left| W_i\left(f,t_0\right)\right|^2df.
\end{equation}

To detect states of sleep and awake the time dependences of the integral energy  $\langle E^{\Delta f}_i(t) \rangle$~(\ref{eq:energy}) in bands $\mathbf{\Delta f} = \left\{\left[2.5; 4.5\right], \left[5; 10\right], \left[10.5; 12.5\right], \left[15; 18\right]\right\}$~Hz. At the first stage, individual threshold values $\mathrm{Tr}_{1,2,3}$ were detected for the ECoG records of each animal. At the second stage, in a $1/2$~s time window, comparisons were made of the current energy characteristics of oscillatory processes at $\left\{{\rm{ECoG}}_1(t), {\rm{ECoG}}_2(t), {\rm{ECoG}}_3(t) \right\}$  for bands $\mathbf{\Delta f}$ with threshold values  $\mathrm{Tr}_{1,2,3}$. The start and end times of NREM sleep states were determined by the method of double crossings of various thresholds $\mathrm{Tr}_{1,2}$. Using of this method is illustrated in Fig.~1(f). The threshold $\mathrm{Tr}_{3}$  was needed for detection of peak-wave discharges in rats with epileptic activity.

The agreement between the results of wavelet analysis and the expert assessments including additional analysis of video recordings and myography, was quite high reaching sensitivity values of $[92; 99]$~\%. The automatic analysis of one animal's hourly record requires about 10~--~15 minutes if this method is used with parallelization of calculations.

\section{How $\mu$ and $\sigma$ calculated by ECoGs relates to Behavioral Sleep?}
\label{sec:discuss_mu_sigma}

Manual scoring of PSG remains the golden standard in sleep-waking cycle evaluation in rats. PSG usually includes two ECoG channels and an electromyography EMG channel (neck muscles activity).
 The first ECoG channel displays the general cortical activity, while the second one is responsible for the hippocampal activity.
 The hippocampal theta-rhythm is one of REM sleep hallmarks (electrode is implanted over parietal or anterior occipital cortex). 
 A PSG period with low amplitude activity in ECoG channels and EMG activity exceeding particular threshold value (conditionally specified for each animal) is classified as wakefulness. High amplitude activity in ECoG channels with delta frequency predominance ($0.1-4$~Hz), sleep spindles periodical appearance (specific patterns in sleep ECoG with $12-16$~Hz frequency) and low amplitude of muscles activity together speak of NREM sleep. 
 During REM sleep EMG signal amplitude  is minimal  and both ECoG channels demonstrate low amplitude activity with prominent theta-rhythm in the ``hippocampal'' channel ($4-8$~Hz).

Behavioral sleep includes NREM and REM sleep. 
 During NREM sleep delta activity of thalamo-cortical network synchronizes cortical neurons firing. This synchronization leads to synchronous rise of signals amplitude in different ECoG channels. 
 Further one can see simultaneous increase of computed means and standard deviations values for these channels.  
 After either REM sleep or wakefulness onset the reverse process takes place, and there is a simultaneous decrease of means and standard deviations values computed for ECoG channels. 
 Also signal amplitude in the ``hippocampal'' ECoG channel often appears to be more stable during REM sleep in comparison to wakefulness (lower vs higher standard deviation). 
 This could additionally facilitate the behavioral sleep and wakefulness discrimination and could possibly fit with a small advantage of using a pair ($\mathrm{ECoG}_2$,$\mathrm{ECoG}_3$) for it. 
 It is also worth noting that REM sleep in rats is quite rare during the dark 12-h period for which our analysis has been performed (see for instance~\textit{Simasko and Mukherjee}~\cite{simasko2009novel}). 
 Thus, the accuracy of our scoring method shouldn’t be greatly reduced in case when REM sleep is wrongly defined by ANN as wakefulness.

\section{Ground-truth data}
\label{sec:discuss_gtd}

It is worth noting that the primary partitioning of the ECoG recordings into WS and BS fragments was performed using an algorithm based on a continuous wavelet transform (CWT, \cite{runnova2021}).
 Therefore, comparing the accuracy of this algorithm with other self-sustaining approaches is not quite correct. 
 The algorithm suggested in this paper should be seen primarily as a computationally simple  technique capable of replacing the CWT, which requires significant computational resources.

As it is shown in \cite{runnova2021} the quality $Q$ of automatic wavelet-based assessment of behavioral sleep in comparison with manual detection is in range $84-98\%$ with median at $91\%$.
 One of the following tasks is to train and test our ANN algorithm on the expert markup of ECoG signals obtained with wavelet-based method and to compare them for other animals excluded during training.

\section{Data normalization problem}
\label{sec:discuss_normalization}

As noted in Sect.~1.3.1, ECoG implementations are first normalized according to Eq.~(1).
 This leads to stabilizing the deviation of local mean calculated in shifting time window of ECoG signal. 
 Next, the set of $\mu_i(t_j)$ (as well as $\sigma_i(t_j)$) values becomes the subjects of another normalization, which is necessary to compress its values in $[0;1]$ range before using them in ANN.

The first ECoG normalization~(1) is a ``bottle neck'' of our algorithm universality.
 Our studies on experimental animals showed that the ANN trained on one of them provides acceptable accuracy with the others (whose data were not used in the training process, see Fig.~2). However, we still cannot claim that the algorithm will provide similar accuracy for the records of other animals.
 Individual characteristics of some animals, errors in the installation of electrodes, usage of electrodes of a different manufacturer can potentially introduce significant changes in the statistical characteristics of the signal. 
 These changes may not be compensated by the normalization~(1) and the algorithm error can be therefore increased significantly.

This problem can be solved by using other approaches to normalize the ECoG signals. 
 So far, we have conducted a number of experiments using normalizations based not on dividing the signal of each channel by its average value, but on the ratio of signals of different ECoG channels (potentially, this approach could eliminate or significantly reduce the influence of the signal spread).
 However, the use of such dividing unfortunately reduces the accuracy to 60\%. 
 The search for an optimal and universal method of data normalization is a topic for further research and discussion. 

\section{Window size and offset}
\label{sec:discuss_window}

The duration of an averaging window $\delta = 10$~sec is motivated by the further reasons. 
 The duration of BS in rats is on the order of tens or hundreds of seconds.
 Since the choice of $\delta$ determines the temporal resolution of our classifier, the largest window duration should not exceed the order of 10 seconds.
 On the other hand, very short $\delta$ does not provide good calculation of statistical characteristics.
 The time window duration in order 1 second contains not so many points (400 in case $f_d=400$~samples in second) and mean value can depend on phase of slow wave oscillation.
 Therefore, taking into account the above limitations of the $\delta$ range, the duration of 10 seconds was chosen empirically as providing the maximum accuracy of the classifier.

\begin{figure}[h!]
\center{\includegraphics[width=0.75\linewidth,keepaspectratio]{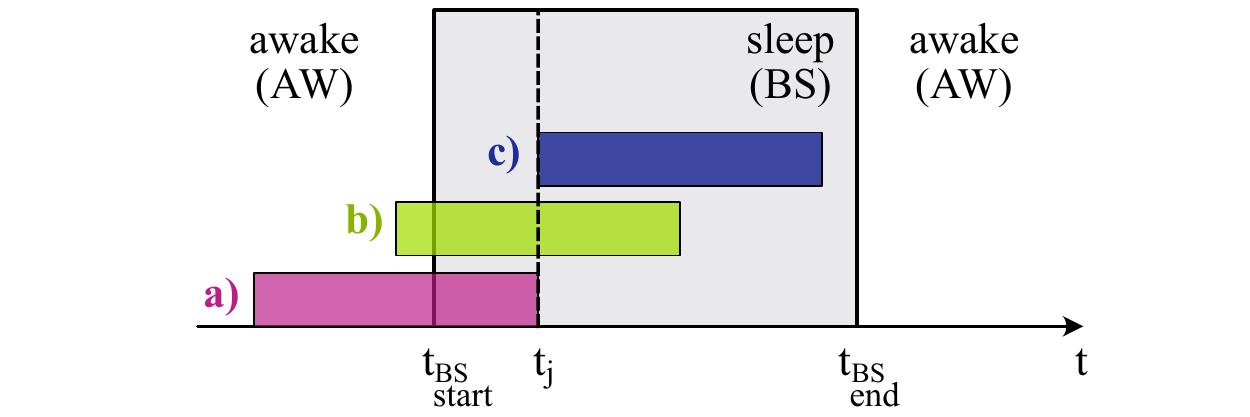}}
\caption{Variants of time window offset. a) from $t_j-\delta$ to $t_j$; b) from $t_j-\delta/2$ to  $t_j+\delta/2$; c) from $t_j$ to $t_j+\delta$. AW and BS mark-up here obtained via CWT.}
\label{fig:window}
\end{figure}

The another subject matter is the starting point for an averaging window.
 The wavelet based approach gives the markup with a one-to-one correspondence of time and the appearance of BS $t=t_{BS start}$. 
 However, in order to obtain the mean value and standard deviation, it is necessary to calculate these characteristics in a certain window. 
 Therefore, it is of particular interest to understand how the beginning of the averaging window correlates with the time of initial markup $t_{BS start}$. 
 This is shown in more detail in Fig.~\ref{fig:window},a).
 If the BS fragment began at some point $t_{BS start}<t_j$ and $t_j - t_{BS start}<\delta$, then the part of time window with duration $\delta-(t_{BS start}-t_j)$ lies already in a WS part. 
 In the case shown in Fig.~\ref{fig:window},c) we can obtain similar situation, when BS fragment ends before the time window end.
 Then situation in Fig.~\ref{fig:window},b) looks more correct because the time window cannot go beyond the BS fragment by more than $\delta/2$.
 However, the results of ANN's training show that the greatest accuracy is ensured when the time interval from $(t_j-\delta)$ to $t_j$  corresponds to time moment $t=t_j$ on markup, as it shown in Fig.~\ref{fig:window},a).
 Presumably, this can be explained by the fact that our method based on mean and standard deviation of ECoG is very sensitive to the beginning of the averaging window and probably even allows to detect BS precursors.


\end{document}